\documentclass[fleqn,10pt]{wlscirep}
\usepackage[utf8]{inputenc}
\usepackage[T1]{fontenc}

\newcommand{\denselist}{\vspace{-3pt} \itemsep -2pt\parsep=-1pt\partopsep -2pt}
\title{Dual-Scale Temporal Fusion Reveals Structured Predictability in Subseasonal-to-Seasonal Temperature Prediction }

\author[1]{Elnaz Bashir}
\author[2]{Jiali Wang}
\author[1,*]{Lin Yan}
\affil[1]{Department of Computer Science, Iowa State University, Ames, IA 50011, USA}
\affil[2]{Environmental Science Division, Argonne National Laboratory, Lemont, IL 60439, USA}

\affil[*]{linyan@iastate.edu}

\keywords{subseasonal-to-seasonal (S2S) prediction $|$ temperature forecasting $|$ dual-scale modeling $|$ horizon-conditioned learning $|$topology-aware modeling}

\begin{abstract}
Subseasonal-to-seasonal (S2S) temperature forecasts, spanning several weeks to a few months, are critically needed in agriculture practice, energy planning, and extreme-weather induced risk management, yet their reliability varies substantially across seasons and regions. Forecast skill is often attributed primarily to lead time, but this perspective does not fully explain the spatiotemporal patterns of predictability. Here we show that S2S predictability is organized across interacting temporal components, spatial heterogeneity, and large-scale pattern coherence, and that this structure can be explicitly characterized and exploited.
We develop a dual-scale learning framework that separates calendar-aligned historical climate context from lead-time matched recent weather evolution, combining them through spatially adaptive fusion to enable stable temperature forecasts across the 30 to 90-day window. The learned fusion weights reveal that the balance between these two temporal scales shifts systematically with season and geography: during winter, interannual context dominates over high latitudes and complex terrain where forecast is the most difficult, while summer predictions reflect a more balanced temporal contribution across the domain. This spatially explicit reorganization of predictability, rather than simple lead-time decay, emerges as the primary determinant of forecast skill within the subseasonal window. Topology-aware structural constraints further improve spatial coherence of predicted temperature fields, stabilizing large-scale pattern organization particularly over complex terrain.
These results reframe S2S predictability as a structured, multi-scale phenomenon, providing a more interpretable foundation for improving forecast systems and informing their use in practice.
\end{abstract}
\begin{document}

\flushbottom
\maketitle
%
%
\thispagestyle{empty}


\section*{Introduction}
Subseasonal-to-seasonal (S2S) prediction is widely recognized as a frontier problem in both scientific and operational practices because it must bridge the gap between medium-range weather forecasting and seasonal outlooks~\cite{WhiteDomeisenAcharya2022,BeckerKirtmanLHeureux2022}. At these lead times, forecast skill inherited from initial conditions decays rapidly, while boundary-forced signals and slowly varying modes have not yet emerged as robust and locally actionable sources of predictability~\cite{Hoskins2013,VitartRobertson2018,Vitart2017,RobertsonVitartCamargo2020,RobertsonVitart2018}.

The importance of S2S  prediction extends beyond theoretical interest.
Many high-consequence societal decisions are made on a 4–12 week planning horizon, where day-to-day weather forecasts are no longer sufficient and seasonal averages are too coarse to support action.
In agriculture and food systems, decisions such as planting and harvest timing, fertilization, and irrigation scheduling often require actionable outlooks beyond two weeks~\cite{KlemmMcPherson2018, WhiteDomeisenAcharya2022}. 
In water management, reservoir operations and allocation planning similarly rely on S2S guidance to anticipate drought and unusual demand~\cite{HwangOrensteinCohen2019, SenguptaSinghDeFlorio2022, WhiteDomeisenAcharya2022}. 
Energy demand planning, heat-risk preparedness, and compound-risk management also benefit from skillful temperature outlooks on the S2S horizon~\cite{ZscheischlerWestraVan-Den-Hurk2018, PendergrassMeehlPulwarty2020}.

Despite decades of progress in numerical weather prediction, deterministic skill at S2S lead times remains limited, and uncertainty quantification typically relies on large ensemble systems~\cite{LeutbecherPalmer2008, PalmerBuizzaDoblas-Reyes2009, RobertsonVitartCamargo2020}. 
To accelerate progress on this timescale, the S2S Prediction Project, led by the World Weather Research Programme (WWRP) and the World Climate Research Programme (WCRP), assembled a multi-center database of near-real-time ensemble forecasts and reforecasts extending to about 60 days, enabling more consistent evaluation across models~\cite{VitartRobertson2018,Vitart2017}. Even with this infrastructure, operational skill in near-surface temperature usually declines substantially beyond two weeks and exhibits strong seasonality and spatial dependence, underscoring the challenge of representing multiscale variability and teleconnections as lead time increases. 
These limitations motivate two broad research directions: improving numerical models and developing data-driven or hybrid systems that can better exploit historical information and systematic structure~\cite{LeutbecherPalmer2008, PalmerBuizzaDoblas-Reyes2009, MouatadidOrensteinFlaspohler2023}.

Artificial Intelligence (AI) has recently transformed medium-range weather forecasting. Models such as Pangu-Weather and GraphCast have demonstrated striking gains in multi-variable global prediction out to about 10 days, while benchmark efforts such as WeatherBench, WeatherBench 2, and Extreme Weather Bench have enabled more transparent and reproducible comparison of data-driven methods~\cite{BiXieZhang2023, LamSanchez-GonzalezWillson2023, RaspDuebenScher2020, RaspHoyerMerose2024, EWB2024}. These advances have stimulated growing interest in extending AI-based prediction beyond the two-week horizon. FuXi-S2S reported global daily-mean forecasts out to roughly 42 days for a broad suite of variables, representing one of the clearest demonstrations of AI competitiveness at the S2S scale~\cite{ChenZhongLi2024}. More recent diffusion-based approaches have also shown progress toward stable 70-90-day forecasts in large-scale generative settings, highlighting the rapid evolution of AI for S2S prediction, although these systems remain new and require careful interpretation with respect to task design and evaluation protocols~\cite{HatanpaaKuStock2025, StockArcomanoKotamarthi2025}. 


One reason is that skill degradation is often framed primarily as a function of forecast horizon, that is, as a roughly monotone decay with lead time. This encourages one-dimensional evaluation and can obscure other regime-dependent factors that matter to end users~\cite{RobertsonVitart2018, WhiteDomeisenAcharya2022}. 
However, S2S predictability is modulated not only by season but also by background state and large-scale modes that reorganize error growth and teleconnection pathways. 
For example, the Madden–Julian Oscillation is widely recognized as a major source of subseasonal predictability, with teleconnections that influence extratropical circulation and surface temperature patterns~\cite{WheelerHendon2004, Zhang2005, KimVitartWaliser2018}.
These considerations suggest that S2S temperature skill should be examined not only as a function of lead time, but also through the lens of seasonally varying spatial structure.

\begin{figure}[t]
\centering
\includegraphics[width=14cm]{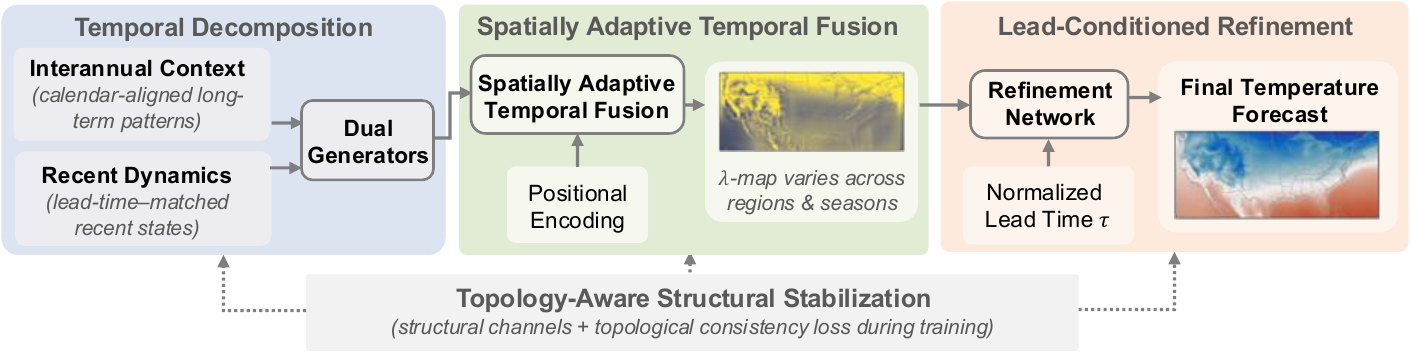}
\caption{Conceptual overview of a dual-scale  S2S temperature forecasting architecture.
Historical climate context and recent weather evolution are processed by parallel generators and combined through spatially adaptive temporal fusion, producing regionally varying fusion weights. A lead-conditioned refinement network improves coherence of the fused forecast. Topology-aware structural descriptors are incorporated within the network, with additional consistency constraints applied during optimization.}
\label{fig:ML}
\end{figure}

In this work, we revisit S2S temperature prediction over the continental United States at 30–90-day lead times from a structural perspective. 
We adopt a dual-scale fusion model that separates calendar-aligned historical climate context from lead-time–matched recent weather evolution and integrates them through spatially adaptive fusion. This design enables explicit analysis of how temporal information is weighted across seasons and geographic regions.
To promote stable spatial organization, we incorporate topology-aware structural descriptors together with consistency constraints that encourage preservation of large-scale spatial coherence.
The overall architecture is illustrated in Fig.~\ref{fig:ML}.

Here, we use the term structured predictability to refer to the explicit representation of temporal decomposition, spatial heterogeneity, and geometric coherence in S2S temperature forecasting. Temporally, temperature evolution is represented through separable components operating on distinct scales, including interannual context versus recent day-to-day evolution. Spatially, predictability weighting varies across regions. Geometrically, large-scale pattern coherence is encouraged through topology-aware constraints. This formulation moves beyond uniform horizon-based modeling by explicitly accounting for multi-scale temporal behavior, region-dependent skill variation, and spatial dependency structure.

Our analysis yields three main findings. First, within the 30–90-day window, forecast errors exhibit a pronounced seasonal organization: winter errors are systematically larger and more spatially heterogeneous than summer errors, whereas within-season differences across lead times are comparatively modest. 
Second, the learned fusion weights adapt to local forecast difficulty in a season-dependent manner, indicating that the balance between recent weather evolution and long-term climate context shifts across forecasting regimes. 
Third, topology-aware structural information improves spatial coherence and accelerates convergence, providing a practical mechanism for stabilizing learning without altering the central adaptive fusion behavior.



\begin{figure*}[t!]
\centering
\includegraphics[width=\linewidth]{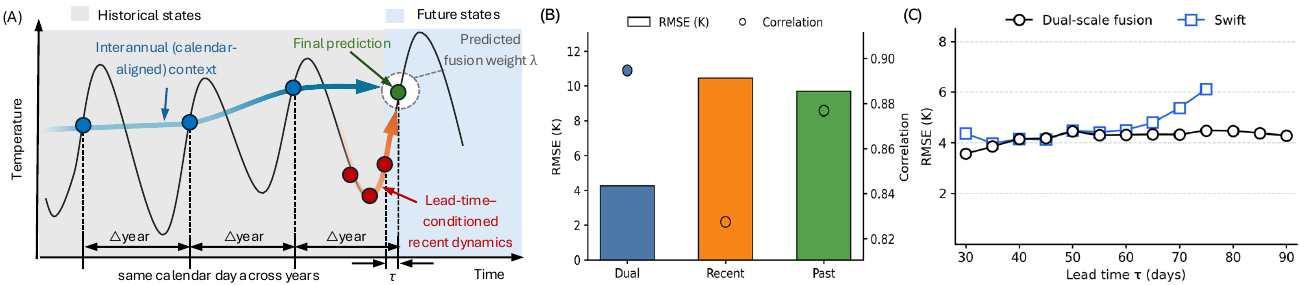}
\caption{Dual-scale formulation and forecast characteristics.
(A) Conceptual schematic illustration of the dual-scale fusion framework, combining calendar-aligned interannual context and lead-time–matched recent dynamics. A spatially adaptive fusion weight $\lambda$ is predicted by the model to modulate their relative contributions and produce the final forecast at lead time $\tau$.
(B) Forecast error and spatial correlation metrics for the dual-scale formulation and corresponding single-scale variants. RMSEs and correlations are averaged over lead times from 30 to 90 days sampled at 5-day intervals.
(C) Forecast skill of our dual-scale fusion model across lead times $\tau$ over the held-out test period, compared with an external S2S stochastic AI foundation forecasting baseline (Swift~\cite{StockArcomanoKotamarthi2025}). Swift results are available up to 75 days and are computed from ensemble predictions (32 initial times, 15 members each). 
}
\label{fig:Dual}
\end{figure*}

\section*{Results}

\begin{figure*}[t!]
\centering
\includegraphics[scale=0.6]{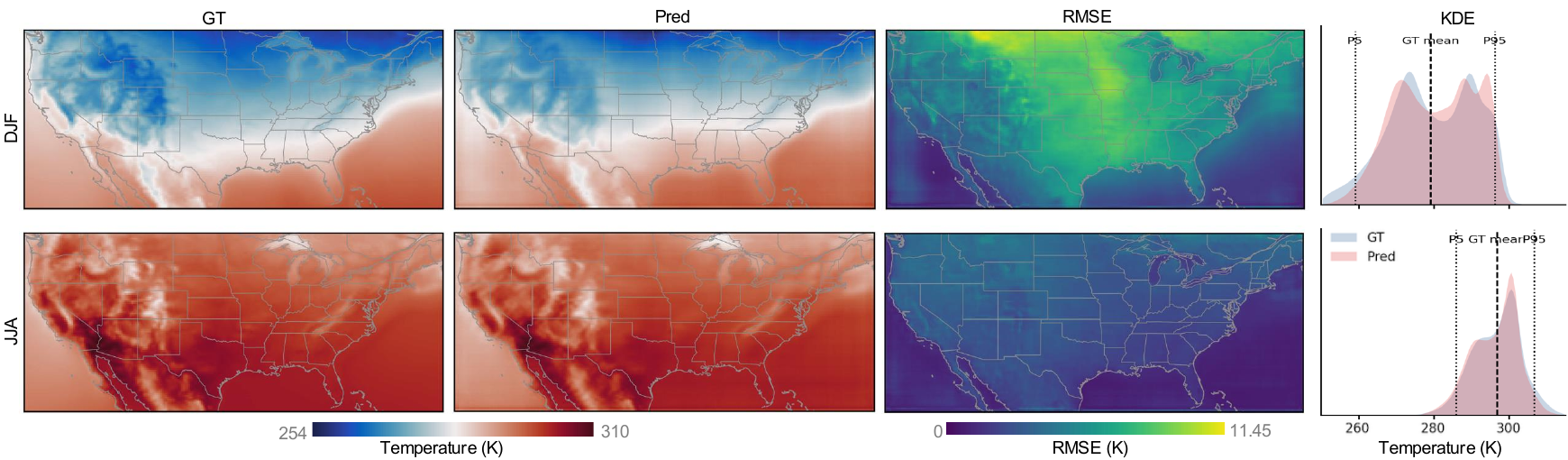}
\caption{Seasonal spatial structure and distributional consistency of S2S temperature prediction.
Seasonal mean near-surface temperature over CONUS comparing ground truth (GT), model prediction (Pred), and spatial root-mean-square error (RMSE) averaged across lead times from 30–90 days (5-day intervals). Winter (DJF) and summer (JJA) are shown in separate rows. GT and Pred share identical color scales within each season to facilitate direct structural comparison, while RMSE is shown using a fixed scale across seasons. 
Kernel density estimates (right) summarize the temperature distributions and illustrate the distributional agreement between GT and Pred. The dashed line indicates the GT mean, and the dotted lines mark the 5th and 95th percentiles of GT. The overlap ratios between the GT and Pred tails are 0.48 for values below P5 and 0.69 for values above P95 in winter, and 0.99 and 0.70, respectively, in summer. Despite increased intrinsic variability in winter, the model preserves both large-scale spatial organization and bulk distributional characteristics across seasons. Seasonal summary statistics are reported in Table S1.}
\label{fig:Seasonal}
\end{figure*}

\begin{figure}[t]
\centering
\includegraphics[width=11.4cm]{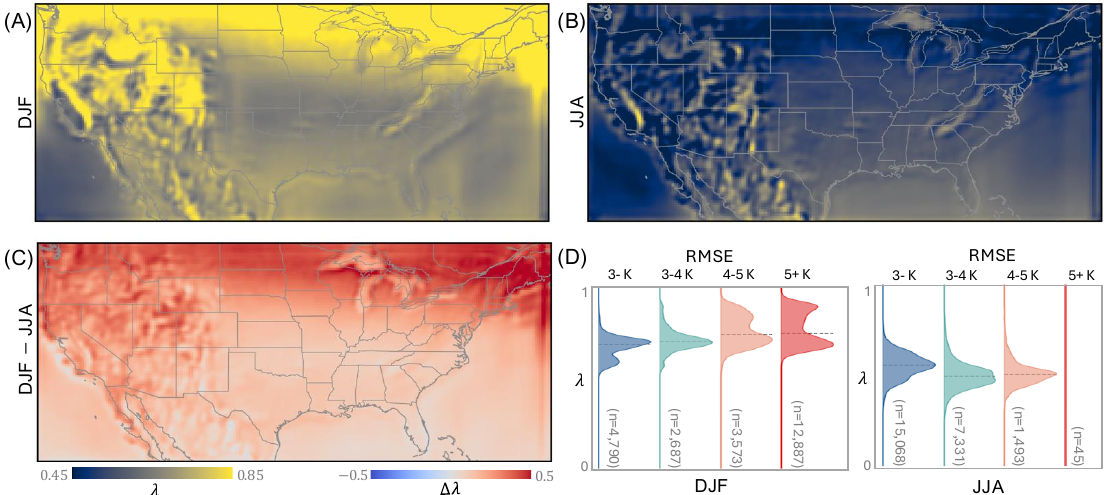}
\caption{
Seasonal reorganization of the learned fusion weight $\lambda$ and its dependence on predictability regimes.
(A) Spatial distribution of $\lambda$ during winter (DJF).
(B) Spatial distribution of $\lambda$ during summer (JJA).
(C) Seasonal difference, $\Delta\lambda = \lambda_{\mathrm{winter}} - \lambda_{\mathrm{summer}}$.
Winter exhibits enhanced $\lambda$ over inland and higher-latitude regions, indicating increased reliance on interannual variability under reduced subseasonal predictability.
(D) Distributions of $\lambda$ stratified by seasonal mean RMSE bins (3- K, 3--4 K, 4--5 K, 5+ K) for DJF (left) and JJA (right).
In winter, $\lambda$ shifts systematically toward higher values with increasing RMSE, whereas summer shows substantially weaker regime-dependent modulation.
These results indicate that $\lambda$ encodes seasonally structured predictability organization rather than reflecting direct pixelwise error sensitivity.
}
\label{fig:lambda}
\end{figure}

Building on the dual-scale architecture outlined in Fig.~\ref{fig:ML}, we first examine its temporal structure (Fig.~\ref{fig:Dual}), then its seasonal and spatial manifestations of predictability (Fig.~\ref{fig:Seasonal} and Fig.~\ref{fig:lambda}), and finally the role of topology-aware cues in stabilizing spatial coherence (Fig.~\ref{fig:topology}).




\subsection*{Dual-scale formulation and temporal stability}
Fig.~\ref{fig:Dual} highlights the temporal decomposition at the core of the dual-scale fusion framework.
At each grid point, temperature evolution is represented as the combination of a recurring annual background pattern aligned across years and lead-time-matched information from recent weather states.
Rather than blending these influences implicitly, the framework represents them as complementary temperature estimates derived from historical climate context and recent weather evolution(Fig.~\ref{fig:Dual}A).
These two components are combined through a spatially adaptive fusion weight, $\lambda$, predicted from the input fields.
Unlike fixed or spatially uniform combinations, this design allows the relative contribution of long-term climate context and recent weather evolution to vary across regions, making $\lambda$ a spatially explicit indicator of how temporal information is weighted.

To evaluate the benefit of temporal decomposition, we compare the dual-scale fusion framework with two single-scale variants that rely exclusively on either historical climate context or recent weather evolution. 
As shown in Fig.~\ref{fig:Dual}B, combining both components consistently reduces forecast error and improves spatial correlation relative to either single-scale variant, indicating that the two timescales provide complementary predictive information.

To further contextualize model performance, we compare our results with those from an external, stochastic AI foundation model, Swift~\cite{StockArcomanoKotamarthi2025} in Fig.~\ref{fig:Dual}C. Swift provides forecasts up to 75 days at a spatial resolution of about 150 km. 
To ensure a fair comparison, our predictions were downsampled to match this resolution prior to evaluation.
Within the overlapping lead-time range (approximately 30–70 days), our dual-scale fusion model shows comparable performance while maintaining relatively stable error behavior across the full 30–90-day window.
Representative comparisons are provided in Fig. S3.

\subsection*{Seasonal prediction quality and spatial–distributional structure}
Fig.~\ref{fig:Seasonal} contrasts representative winter (DJF) and summer (JJA) forecasts averaged across lead times from 30 to 90 days.
Transitional seasons are shown in Fig. S4. Seasonal statistics are summarized in Table S1, while Table S2 reports seasonal contrast and lead-time sensitivity relative to a simple day-of-year climatology baseline.
Pronounced seasonal differences emerge in both spatial structure and error magnitude.

In DJF, the ground-truth temperature field exhibits sharper meridional gradients and more intricate spatial texture. 
The predicted field preserves the large-scale organization but displays amplified errors across high latitudes and complex terrain. 
In contrast, the JJA ground-truth temperature fields are smoother and more spatially coherent, and the predictions align closely with regional temperature patterns, especially terrain-driven localized gradients. Mean RMSE decreases from approximately 5.8 K in DJF to 2.7 K in JJA, highlighting a strong seasonal contrast that exceeds variation across lead times within the same 30–90-day window.

Spatial error maps indicate that this contrast is systematic rather than localized. Winter errors extend across broad continental regions, whereas summer errors are smaller in magnitude and confined mainly to areas with strong spatial gradients (e.g., mountains). The difference, therefore, reflects a shift in the overall error structure rather than a few isolated hotspots.

Distributional analysis reinforces this picture. 
Kernel density estimates (KDEs) show substantial overlap between predicted and ground truth temperatures in both seasons, with an overlap ratio of 0.95 and 0.93 in summer and winter, respectively; however, winter distributions exhibit broader tails and greater spread. The seasonal contrast thus arises predominantly from differences in variability rather than from systematic mean bias. 

This seasonal contrast remains evident across lead times, with the winter–summer separation exceeding within-season variation over the 30–90-day window (Fig.~S5A). This indicates that, within the subseasonal range, seasonal structure exerts a stronger influence on forecast quality than horizon-dependent degradation~\cite{DomeisenButlerCharlton-Perez2020}. This seasonal organization motivates the next analysis of how adaptive temporal fusion responds across seasons and geographic regions.



\subsection*{Seasonal contrasts and space-aware adaptation of the predicted fusion weight}
Having established that seasonal differences dominate lead-time effects in forecast error, we next examine how the fusion weight $\lambda$ varies across space. 
Fig.~\ref{fig:lambda}A–C contrasts winter (DJF) and summer (JJA) $\lambda$ fields and their difference ($\Delta \lambda = \lambda_{\mathrm{winter}} - \lambda_{\mathrm{summer}}$), revealing pronounced and spatially coherent seasonal contrasts. Larger $\lambda$ values correspond to greater reliance on historical climate context, whereas smaller values place more weight on recent weather evolution.

During winter, $\lambda$ is systematically elevated across large portions of the continental United States, particularly over the northern U.S. and mountainous regions, suggesting greater reliance on interannual context under more variable midlatitude environment. In summer, $\lambda$ is lower and more spatially homogeneous, remaining near 0.5 across much of the domain, indicating a more balanced contribution from historical context and recent weather. The difference map (Fig.~\ref{fig:lambda}C) shows that regions with amplified winter $\lambda$ largely coincide with regions of increased forecast error (Fig.~\ref{fig:Seasonal}), indicating that the fusion strength intensifies where predictability is lower rather than shifting uniformly with season.

Beyond these seasonal contrasts, elevated $\lambda$ values are also consistently observed over complex terrain, including during summer when forecast errors remain relatively low. This indicates that higher $\lambda$ does not uniformly correspond to larger errors, but instead reflects region-dependent temporal weighting.
For example, during summer, regions with elevated $\lambda$, particularly over mountainous areas, still exhibit relatively low RMSE (Fig.~\ref{fig:Seasonal}). 
These patterns suggest that the fusion mechanism adapts differently across spatial and seasonal conditions, reflecting the interplay between temporal variability and geographic structure rather than a simple monotonic relationship between $\lambda$ and forecast error.

Error-stratified analysis (Fig.~\ref{fig:lambda}D; Fig.~S6; Table~S3) further clarifies this behavior. In winter, mean $\lambda$ increases monotonically across RMSE bins, with higher-error regions assigning greater weight to interannual context. 
This monotonic increase persists across substantial sample sizes, indicating that the pattern is not driven by isolated extremes. 
In summer, by contrast, $\lambda$ remains comparatively stable across error bins, with only weak sensitivity to local RMSE. Transitional seasons display intermediate behavior.
Importantly, $\lambda$ also remains largely stable across forecast horizons within the 30–90-day window (Fig.~S5B), with nearly identical central tendencies at 30, 60, and 90 days. Thus, although forecast error increases modestly with lead time (Fig.~\ref{fig:Dual}C), the fusion weight does not vary systematically with forecast horizon.

Taken together, these results show that adaptive temporal fusion responds primarily to seasonal regime and local forecast difficulty, rather than to lead time alone. In particular, elevated winter $\lambda$ over northern latitudes and complex terrain suggests that the model leans more heavily on historical climate context in settings where recent weather evolution is less reliable. This behavior is physically plausible and consistent with well-known challenges in numerical weather prediction: midlatitude winter is characterized by stronger baroclinic instability and richer mesoscale variability, while the local manifestations of weather, such as snowfall, cold extremes, and wind gusts, are often much harder to predict than the larger-scale synoptic pattern. 
The learned $\lambda$ field, therefore, provides a spatially explicit indicator of when and where interannual context offers a stabilizing benefit relative to recent dynamics. We next examine how topology-aware structure further shapes spatial coherence within this adaptive system.

\begin{figure*}[t!]
\centering
\includegraphics[scale=0.6]{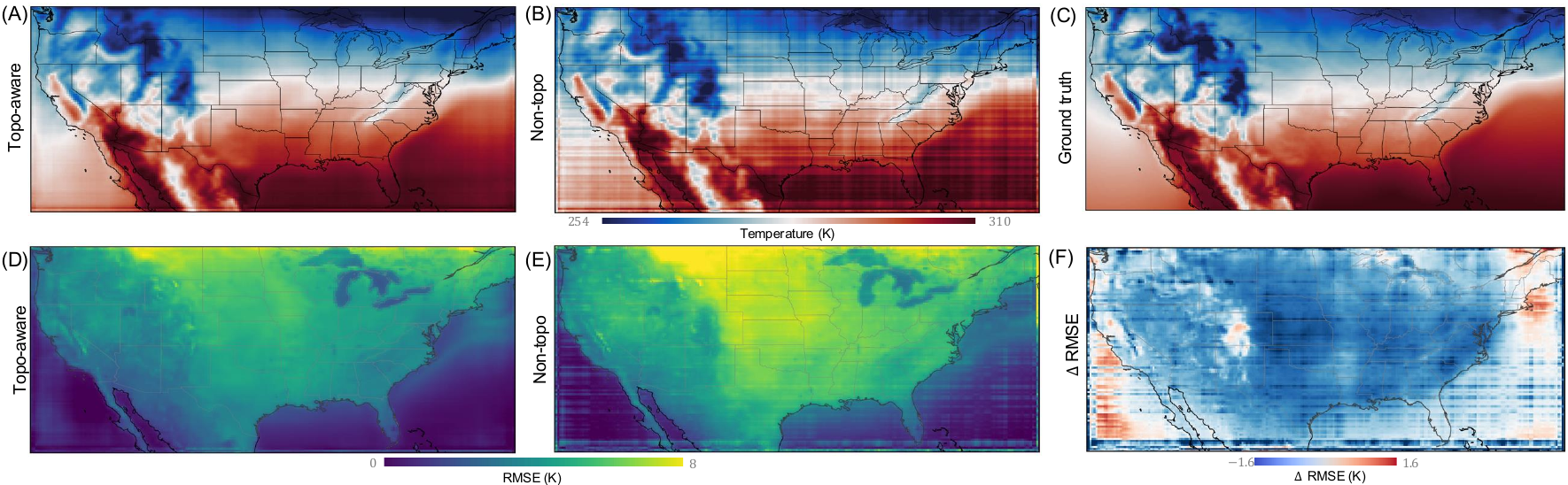}
\caption{
Topology-aware modeling improves spatial coherence and predictive accuracy. The first row (A–C) shows the annual mean temperature fields over the evaluation period, comparing the topology-aware prediction (A), the non-topological baseline (B), and the ground truth (C). The second row (D–F) presents the corresponding spatial RMSE for the topology-aware model (D), the baseline model (E), and their difference ($\Delta$RMSE, F).
Both models capture the large-scale temperature distribution, but the non-topological baseline exhibits grid-aligned artifacts and reduced spatial coherence. In contrast, the topology-aware model produces more realistic and physically consistent spatial patterns. The $\Delta$RMSE map indicates broadly distributed error reduction across most regions. 
}

\label{fig:topology}
\end{figure*}


\subsection*{Topology-aware modeling improves spatial coherence and optimization stability}
Fig. \ref{fig:topology} examines the effect of incorporating topology-aware structural descriptors on spatial prediction behavior. 
Spatial RMSE maps averaged across lead times from 30 to 90 days reveal clear differences between the topology-aware model (Fig. \ref{fig:topology}D) and the matched baseline without topological inputs (Fig. \ref{fig:topology}E). Although the overall reduction in RMSE is moderate, the spatial organization of the error field changes noticeably.

The baseline exhibits visible grid-aligned banding patterns across portions of the domain, whereas the topology-aware formulation produces more realistic spatial patterns and generally smaller errors. In particular, regional gradient structures become more clearly resolved, and the predicted fields align more closely with large-scale terrain-related temperature variations.
Fig. \ref{fig:topology}F further shows that error reductions are broadly distributed rather than confined to isolated regions, indicating that the topology-aware representation improves spatial consistency of predicted temperature fields.

Importantly, incorporating topological information also improves training behavior and predictive consistency (Fig.~S7). These results align with the notion of structured predictability introduced earlier: by helping preserve large-scale spatial coherence while interacting with temporally decomposed inputs and season- and region-dependent fusion weighting, the topology-aware formulation promotes more stable learning of spatial dependencies and more consistent forecast behavior.

\section*{Discussion}
Our study complements physically based understanding of S2S variability and highlights the potential role of data-driven models as diagnostic tools for identifying where and when different sources of predictability become most influential~\cite{BiXieZhang2023}. In particular, the results reveal a structured organization of predictability that can be examined through the proposed dual-scale fusion framework.

\subsection*{A structured view of S2S predictability}
This structured behavior is reflected in three main aspects. First, forecast errors exhibit a clear seasonal contrast, with the largest errors occurring in winter. Winter is characterized by broader error distributions and stronger spatial heterogeneity, particularly over northern and mountainous regions, whereas summer forecasts remain comparatively stable and spatially consistent.  This pattern aligns with the stronger variability and reduced predictability of midlatitude winter conditions, and is further corroborated by consistent improvements over a day-of-year climatology baseline (the historical mean temperature for each calendar day, a standard reference in S2S forecasting~\cite{WhiteDomeisenAcharya2022,VitartRobertson2018}): RMSEs are reduced by 2.26 K in DJF and 3.76 K in JJA (Table S2), suggesting that the observed seasonal contrast reflects physically meaningful structure rather than model-specific artifacts.

Second, the learned fusion weights show systematic spatial and seasonal variation. Higher $\lambda$ values are associated with regions of larger forecast error in winter  (e.g., the northern U.S. and complex terrain), while lower values are observed under more stable summer conditions. Elevated $\lambda$ values also appear consistently over mountainous regions, indicating spatially heterogeneous temporal weighting. Notably, higher $\lambda$ does not uniformly correspond to larger errors, suggesting that the fusion mechanism adapts to both seasonal variability and geographic structure. This relationship should be interpreted as an empirical association rather than a causal mechanism.

Third, incorporating topology-aware structural information improves the coherence of predicted temperature fields. Compared with models without structural constraints, the resulting predictions exhibit more consistent large-scale spatial organization while preserving localized variability, suggesting that topology-aware cues act as an effective inductive bias for stabilizing spatial dependencies.

\subsection*{Implications, Limitations, and Future Directions}

These findings provide a complementary perspective on how S2S predictability is structured, with implications for both interpretation and model design. 
From a modeling perspective, future AI-based approaches may benefit from incorporating multi-scale structure, in which slowly varying background conditions help stabilize longer-range predictions.
From an interpretability perspective, the learned fusion weights provide a more interpretable view of data-driven model behavior by revealing how multi-scale information is utilized across seasons and regions.

Several limitations should be noted. First, while the learned fusion behavior exhibits physically plausible patterns, its relationship to established climate modes and large-scale circulation features has not been explicitly quantified. Future work could investigate connections between the learned representations and known sources of subseasonal predictability, such as teleconnection patterns.

Second, direct quantitative comparison with other data-driven S2S systems is constrained by the limited public availability of model outputs over the full 30–90-day window. Most existing approaches, including FuXi-S2S~\cite{ChenZhongLi2024}, report results up to approximately 42 days, leaving the extended subseasonal range without established AI reference points. Swift represents the closest available comparison; to ensure a fair evaluation, our predictions were coarsened to match Swift's~\cite{StockArcomanoKotamarthi2025} spatial resolution (approximately 150 km) prior to comparison. Establishing standardized benchmarks for the full subseasonal window remains an open challenge for the community, and future evaluation against operational numerical ensemble systems such as ECMWF S2S reforecasts would provide a more complete picture of relative skill.

Third, this study relies primarily on near-surface air temperature as the input variable. Extending the current framework to incorporate additional atmospheric variables could further improve temperature prediction by providing complementary information beyond temperature alone. It may also enable extensions to more challenging settings, such as prolonged extreme events (e.g., heatwaves), where higher temporal resolution and richer inputs are needed to capture evolving dynamics.

Fourth, this framework is developed for a single-variable setting and a regional domain. Extending this approach toward a more comprehensive Earth system setting, in which multiple atmospheric variables are jointly modeled and predicted, would enable a more complete characterization of S2S predictability. In particular, the framework may be useful for variables governed by multi-scale temporal processes, such as components of the water cycle (e.g., precipitation, soil moisture, and evaporation), where different memory timescales and responses to atmospheric forcing interact. In such settings, separating slowly varying background influences from recent forcing may provide further insight into how predictability is organized.

More broadly, this perspective suggests a potential pathway for developing more integrated S2S forecasting systems, in which learning is organized around structured variability rather than forecast lead time alone, potentially improving both interpretability and forecast stability.

\section*{Methods}

This section summarizes the datasets, forecasting formulation, and modeling framework used to analyze S2S temperature predictability. Detailed preprocessing procedures, architectural diagrams, and training specifications are provided in Supporting Information.

\subsection*{Data}

We use the European Centre for Medium-Range Weather Forecasts phase 5 reanalysis dataset (ERA5)~\cite{HersbachBellBerrisford2020}, which provides globally consistent atmospheric fields generated through advanced data assimilation techniques~\cite{DeeUppalaSimmons2011} and has been widely adopted in ML model training for weather predictability~\cite{RaspDuebenScher2020,BiXieZhang2023, LamSanchez-GonzalezWillson2023}.
Our analysis focuses on daily mean 2-m air temperature over the continental United States (CONUS). Hourly ERA5 fields are aggregated to daily means and cropped to a $101\times 237$ spatial region covering the CONUS domain  across all experiments .
The dataset spans 2010--2024, with the first 10 years (2010--2019) for training and the last 5 years (2020--2024) for evaluation. 

\subsection*{Problem formulation and forecasting task}

We consider S2S prediction of daily mean near-surface air temperature at lead times ranging from 30 to 90 days, a window that bridges the scales between medium-range weather forecasting and seasonal forecasting ~\cite{RobertsonVitart2018, WhiteDomeisenAcharya2022}. 

Forecasts across the 30–90-day range are trained within a unified model rather than as separate horizon-specific networks. Lead time $\tau$ is incorporated as a conditioning variable within the refinement module, enabling horizon-dependent adjustment through shared parameters.

The dual-scale fusion mechanism, including the predicted $\lambda$-map, does not receive $\tau$ as an explicit input. However, lead time is implicitly encoded through the horizon-conditioned recent context, whose temporal sampling depends on $\tau$. This design allows the model to learn both common structures (e.g., temporal and spatial organization) and time horizon-dependent variations, while preserving the interpretability of the fusion weight.  Training samples are organized across lead-time groups spanning the 30–90-day window, without imposing globally fixed temporal combinations~\cite{RaspDuebenScher2020,ChenZhongLi2024}.

At S2S scales, predictability reflects contributions from both slowly varying background climate conditions and more rapidly evolving recent atmospheric variability~\cite{Hoskins2013, Vitart2017, KimVitartWaliser2018}. To reflect these distinct temporal influences, model inputs are organized into two complementary contexts:

\begin{itemize}\denselist
    \item \emph{Historical climate context:} calendar-aligned temperature fields from the same day in previous years, capturing climatological baselines and low-frequency variability.
    \item \emph{Recent weather evolution:} horizon-conditioned observations immediately preceding forecast initialization, capturing short-term weather evolution relevant to the target horizon.
\end{itemize}

This separation is calendar-aligned and horizon-conditioned rather than a conventional seasonal decomposition. The two contexts encode distinct predictive signals and are not assumed to contribute equally across space, season, or forecast horizon.

\subsection*{Dual-scale learning framework}
Building on the dual-scale architecture summarized in Fig.~\ref{fig:ML}, we implement the temporal decomposition through two parallel predictive branches that process historical and recent weather contexts separately (detailed in Fig.~S2). Each branch produces a candidate temperature forecast reflecting the structure emphasized by its respective temporal input.
Both branches are implemented as U-Net–style encoder–decoder networks with skip connections, trained with auxiliary adversarial objectives.

Rather than prescribing a fixed or globally uniform temporal weighting, the two candidate forecasts are combined through a spatially resolved fusion $\lambda$-map predicted conditionally on the input fields. Unlike scalar or globally shared gating mechanisms, $\lambda$ is learned as a dense map at the same spatial resolution as the output temperature field. This design enables location-specific modulation of interannual and recent weather contributions.
By explicitly predicting a spatially resolved fusion field, the formulation treats temporal fusion as an interpretable spatial representation rather than an internal network operation. The $\lambda$-map functions as a learned gating variable rather than a physical quantity, but its explicit representation allows post hoc analysis of how temporal information is balanced across space and season.
The fused prediction is subsequently processed by a residual refinement network~\cite{HeZhangRen2016}, which improves spatial coherence while preserving large-scale structure.

\subsection*{Lead-conditioned refinement}

Forecast lead time $\tau$ is encoded as a normalized scalar and broadcast spatially as a conditioning map. This lead-conditioned representation is injected into intermediate layers of a U-Net–based refinement module~\cite{RonnebergerFischerBrox2015}, allowing the model to adjust its residual correction according to forecast horizon.

This refinement stage performs a single learned update rather than iterative diffusion. Its role is to reduce noise and enhance spatial consistency without altering the primary dual-scale fusion mechanism.

\subsection*{Topology-aware structural constraint}
Predictions from data-driven models may exhibit spatial artifacts or inconsistent patterns, particularly in regions with complex variability. To mitigate these issues, we incorporate topology-aware structural descriptors derived from the temperature field as auxiliary input channels (see Fig. S1). These descriptors encode critical-point type, scalar magnitude, and saddle-level contour structure, providing explicit information about spatial organization.

Topological summaries provide compact representations of spatial connectivity and feature persistence that are robust to small perturbations~\cite{Yan2025,YanMasoodSridharamurthy2021}. In this setting, topology-aware channels act as an inductive bias that guides how spatial dependencies are organized during learning.

The topology-aware component does not alter the core forecasting architecture or impose explicit physical constraints. It influences spatial representation learning by providing structural cues. While alternative regularization strategies may achieve related stabilization effects, explicitly encoding structural information offers a principled mechanism for improving spatial coherence within the modeling framework considered here.

\subsection*{Training objective}

The model is trained end-to-end using a composite objective that integrates pixel-level reconstruction, structural similarity~\cite{WangBovikSheikh2004}, adversarial consistency~\cite{GoodfellowPouget-AbadieMirza2014}, and structural regularization terms~\cite{EdelsbrunnerHarerothers2008,Cohen-SteinerEdelsbrunnerHarer2005}. The topological regularization component is activated after an initial warm-up period to avoid interfering with early-stage optimization.

\section*{Funding}

This work was supported in part by startup funds provided by Iowa State University. Grant number: Not applicable. Computational resources were also provided by Iowa State University.

\section*{Acknowledgment}
We thank Dr. Jason Stock and Dr. Rao Kotamarthi of Argonne National Laboratory for sharing the Swift forecasting results, which enabled direct comparison with our study.

\section*{Author contributions statement}

E.B. designed and implemented the experiments, conducted data analysis, and prepared the figures.
J.W. provided domain-specific interpretation of temperature prediction results.
L.Y. conceived the study, developed the topology-aware dual-scale modeling framework, and supervised the research.
All authors contributed to the discussion and writing of the manuscript.

\bibliography{SSTemp-main}

@article{MiyatoKataokaKoyama2018,
	author = {Miyato, Takeru and Kataoka, Toshiki and Koyama, Masanori and Yoshida, Yuichi},
	journal = {arXiv preprint arXiv:1802.05957},
	title = {Spectral normalization for generative adversarial networks},
	year = {2018}}

@article{VitartArdilouzeBonet2017,
	author = {Vitart, Frederic and Ardilouze, Constantin and Bonet, Axel and Brookshaw, Anca and Chen, M and Codorean, C and D{\'e}qu{\'e}, M and Ferranti, L and Fucile, E and Fuentes, M and others},
	journal = {Bulletin of the American Meteorological Society},
	number = {1},
	pages = {163--173},
	title = {The subseasonal to seasonal (S2S) prediction project database},
	volume = {98},
	year = {2017}}

@article{HersbachBellBerrisford2020,
	author = {Hersbach, Hans and Bell, Bill and Berrisford, Paul and Hirahara, Shoji and Hor{\'a}nyi, Andr{\'a}s and Mu{\~n}oz-Sabater, Joaqu{\'\i}n and Nicolas, Julien and Peubey, Carole and Radu, Raluca and Schepers, Dinand and others},
	journal = {Quarterly journal of the royal meteorological society},
	number = {730},
	pages = {1999--2049},
	publisher = {Wiley Online Library},
	title = {The ERA5 global reanalysis},
	volume = {146},
	year = {2020}}

@article{WangBovikSheikh2004,
	author = {Wang, Zhou and Bovik, Alan C and Sheikh, Hamid R and Simoncelli, Eero P},
	journal = {IEEE transactions on image processing},
	number = {4},
	pages = {600--612},
	publisher = {IEEE},
	title = {Image quality assessment: from error visibility to structural similarity},
	volume = {13},
	year = {2004}}

@book{EdelsbrunnerHarer2010,
	author = {Edelsbrunner, Herbert and Harer, John},
	publisher = {American Mathematical Soc.},
	title = {Computational topology: an introduction},
	year = {2010}}

@book{Project2025,
	author = {The GUDHI Project},
	edition = {3.11.0},
	publisher = {GUDHI Editorial Board},
	title = {GUDHI User and Reference Manual},
	url = {https://gudhi.inria.fr/doc/3.11.0/},
	year = {2025}}

@article{StockArcomanoKotamarthi2025,
	author = {Stock, Jason and Arcomano, Troy and Kotamarthi, Rao},
	journal = {arXiv preprint arXiv:2509.25631},
	title = {Swift: An autoregressive consistency model for efficient weather forecasting},
	year = {2025}}

@misc{EWB2024,
	howpublished = {https://github.com/brightbandtech/extremeweatherbench},
	title = {Extreme Weather Bench},
	year = {2024}}

@article{DomeisenButlerCharlton-Perez2020,
	author = {Domeisen, Daniela IV and Butler, Amy H and Charlton-Perez, Andrew J and Ayarzag{\"u}ena, Blanca and Baldwin, Mark P and Dunn-Sigouin, Etienne and Furtado, Jason C and Garfinkel, Chaim I and Hitchcock, Peter and Karpechko, Alexey Yu and others},
	journal = {Journal of Geophysical Research: Atmospheres},
	number = {2},
	pages = {e2019JD030920},
	publisher = {Wiley Online Library},
	title = {The role of the stratosphere in subseasonal to seasonal prediction: 1. Predictability of the stratosphere},
	volume = {125},
	year = {2020}}

@inproceedings{Cohen-SteinerEdelsbrunnerHarer2005,
	author = {Cohen-Steiner, David and Edelsbrunner, Herbert and Harer, John},
	booktitle = {Proceedings of the twenty-first annual symposium on Computational geometry},
	pages = {263--271},
	title = {Stability of persistence diagrams},
	year = {2005}}

@article{EdelsbrunnerHarerothers2008,
	author = {Edelsbrunner, Herbert and Harer, John and others},
	journal = {Contemporary mathematics},
	number = {26},
	pages = {257--282},
	publisher = {Providence, RI: American Mathematical Society},
	title = {Persistent homology-a survey},
	volume = {453},
	year = {2008}}

@article{GoodfellowPouget-AbadieMirza2014,
	author = {Goodfellow, Ian J and Pouget-Abadie, Jean and Mirza, Mehdi and Xu, Bing and Warde-Farley, David and Ozair, Sherjil and Courville, Aaron and Bengio, Yoshua},
	journal = {Advances in neural information processing systems},
	title = {Generative adversarial nets},
	volume = {27},
	year = {2014}}

@inproceedings{RonnebergerFischerBrox2015,
	author = {Ronneberger, Olaf and Fischer, Philipp and Brox, Thomas},
	booktitle = {International Conference on Medical image computing and computer-assisted intervention},
	organization = {Springer},
	pages = {234--241},
	title = {U-net: Convolutional networks for biomedical image segmentation},
	year = {2015}}

@inproceedings{HeZhangRen2016,
	author = {He, Kaiming and Zhang, Xiangyu and Ren, Shaoqing and Sun, Jian},
	booktitle = {Proceedings of the IEEE conference on computer vision and pattern recognition},
	pages = {770--778},
	title = {Deep residual learning for image recognition},
	year = {2016}}

@article{DeeUppalaSimmons2011,
	author = {Dee, Dick P and Uppala, SMꎬ and Simmons, Adrian J and Berrisford, Paul and Poli, Paul and Kobayashi, Shinya and Andrae, U and Balmaseda, MA and Balsamo, G and Bauer, d P and others},
	journal = {Quarterly Journal of the royal meteorological society},
	number = {656},
	pages = {553--597},
	publisher = {Wiley Online Library},
	title = {The ERA-Interim reanalysis: Configuration and performance of the data assimilation system},
	volume = {137},
	year = {2011}}

@article{BeckerKirtmanLHeureux2022,
	address = {Boston MA, USA},
	author = {Emily J. Becker and Ben P. Kirtman and Michelle L'Heureux and {\'A}ngel G. Mu{\~n}oz and Kathy Pegion},
	doi = {10.1175/BAMS-D-20-0327.1},
	journal = {Bulletin of the American Meteorological Society},
	number = {3},
	pages = {E973 - E995},
	publisher = {American Meteorological Society},
	title = {A Decade of the North American Multimodel Ensemble (NMME): Research, Application, and Future Directions},
	url = {https://journals.ametsoc.org/view/journals/bams/103/3/BAMS-D-20-0327.1.xml},
	volume = {103},
	year = {2022}}

@article{KlemmMcPherson2018,
	author = {Klemm, Toni and McPherson, Renee A},
	journal = {Journal of Applied Meteorology and Climatology},
	number = {9},
	pages = {2129--2140},
	title = {Assessing decision timing and seasonal climate forecast needs of winter wheat producers in the South-Central United States},
	volume = {57},
	year = {2018}}

@article{ZscheischlerWestraVan-Den-Hurk2018,
	author = {Zscheischler, Jakob and Westra, Seth and Van Den Hurk, Bart JJM and Seneviratne, Sonia I and Ward, Philip J and Pitman, Andy and AghaKouchak, Amir and Bresch, David N and Leonard, Michael and Wahl, Thomas and others},
	journal = {Nature climate change},
	number = {6},
	pages = {469--477},
	publisher = {Nature Publishing Group UK London},
	title = {Future climate risk from compound events},
	volume = {8},
	year = {2018}}

@article{PendergrassMeehlPulwarty2020,
	author = {Pendergrass, Angeline G and Meehl, Gerald A and Pulwarty, Roger and Hobbins, Mike and Hoell, Andrew and AghaKouchak, Amir and Bonfils, C{\'e}line JW and Gallant, Ailie JE and Hoerling, Martin and Hoffmann, David and others},
	journal = {Nature Climate Change},
	number = {3},
	pages = {191--199},
	publisher = {Nature Publishing Group UK London},
	title = {Flash droughts present a new challenge for subseasonal-to-seasonal prediction},
	volume = {10},
	year = {2020}}

@article{Vitart2017,
	author = {Vitart, Fr{\'e}d{\'e}ric},
	journal = {Quarterly Journal of the Royal Meteorological Society},
	number = {706},
	pages = {2210--2220},
	publisher = {Wiley Online Library},
	title = {Madden---Julian Oscillation prediction and teleconnections in the S2S database},
	volume = {143},
	year = {2017}}

@article{KimVitartWaliser2018,
	author = {Kim, Hyemi and Vitart, Fr{\'e}d{\'e}ric and Waliser, Duane E},
	journal = {Journal of Climate},
	number = {23},
	pages = {9425--9443},
	title = {Prediction of the Madden--Julian oscillation: A review},
	volume = {31},
	year = {2018}}

@article{Zhang2005,
	author = {Zhang, Chidong},
	journal = {Reviews of Geophysics},
	number = {2},
	publisher = {Wiley Online Library},
	title = {Madden-julian oscillation},
	volume = {43},
	year = {2005}}

@article{WheelerHendon2004,
	author = {Wheeler, Matthew C and Hendon, Harry H},
	journal = {Monthly weather review},
	number = {8},
	pages = {1917--1932},
	title = {An all-season real-time multivariate MJO index: Development of an index for monitoring and prediction},
	volume = {132},
	year = {2004}}

@article{PalmerBuizzaDoblas-Reyes2009,
	author = {Palmer, Tim N and Buizza, Roberto and Doblas-Reyes, F and Jung, Thomas and Leutbecher, Martin and Shutts, Glenn J and Steinheimer, Martin and Weisheimer, Antje},
	journal = {ECMWF Reading, UK},
	title = {Stochastic parametrization and model uncertainty},
	year = {2009}}

@article{LeutbecherPalmer2008,
	author = {Leutbecher, Martin and Palmer, Tim N},
	journal = {Journal of computational physics},
	number = {7},
	pages = {3515--3539},
	publisher = {Elsevier},
	title = {Ensemble forecasting},
	volume = {227},
	year = {2008}}

@article{Hoskins2013,
	author = {Hoskins, Brian},
	journal = {Quarterly Journal of the Royal Meteorological Society},
	number = {672},
	pages = {573--584},
	publisher = {Wiley Online Library},
	title = {The potential for skill across the range of the seamless weather-climate prediction problem: a stimulus for our science},
	volume = {139},
	year = {2013}}

@article{RaspHoyerMerose2024,
	author = {Rasp, Stephan and Hoyer, Stephan and Merose, Alexander and Langmore, Ian and Battaglia, Peter and Russell, Tyler and Sanchez-Gonzalez, Alvaro and Yang, Vivian and Carver, Rob and Agrawal, Shreya and others},
	journal = {Journal of Advances in Modeling Earth Systems},
	number = {6},
	pages = {e2023MS004019},
	publisher = {Wiley Online Library},
	title = {Weatherbench 2: A benchmark for the next generation of data-driven global weather models},
	volume = {16},
	year = {2024}}

@article{RaspDuebenScher2020,
	author = {Rasp, Stephan and Dueben, Peter D and Scher, Sebastian and Weyn, Jonathan A and Mouatadid, Soukayna and Thuerey, Nils},
	journal = {Journal of Advances in Modeling Earth Systems},
	number = {11},
	pages = {e2020MS002203},
	publisher = {Wiley Online Library},
	title = {WeatherBench: a benchmark data set for data-driven weather forecasting},
	volume = {12},
	year = {2020}}

@inproceedings{HwangOrensteinCohen2019,
	author = {Hwang, Jessica and Orenstein, Paulo and Cohen, Judah and Pfeiffer, Karl and Mackey, Lester},
	booktitle = {Proceedings of the 25th ACM SIGKDD international conference on knowledge discovery \& data mining},
	pages = {2325--2335},
	title = {Improving subseasonal forecasting in the western US with machine learning},
	year = {2019}}

@article{SenguptaSinghDeFlorio2022,
	author = {Sengupta, Agniv and Singh, Bohar and DeFlorio, Michael J and Raymond, Colin and Robertson, Andrew W and Zeng, Xubin and Waliser, Duane E and Jones, Jeanine},
	journal = {Bulletin of the American Meteorological Society},
	number = {10},
	pages = {E2168--E2175},
	title = {Advances in subseasonal to seasonal prediction relevant to water management in the western United States},
	volume = {103},
	year = {2022}}

@article{WhiteDomeisenAcharya2022,
	author = {White, Christopher J and Domeisen, Daniela IV and Acharya, Nachiketa and Adefisan, Elijah A and Anderson, Michael L and Aura, Stella and Balogun, Ahmed A and Bertram, Douglas and Bluhm, Sonia and Brayshaw, David J and others},
	journal = {Bulletin of the American Meteorological Society},
	number = {6},
	pages = {E1448--E1472},
	publisher = {American Meteorological Society},
	title = {Advances in the application and utility of subseasonal-to-seasonal predictions},
	volume = {103},
	year = {2022}}

@book{RobertsonVitart2018,
	author = {Robertson, Andrew and Vitart, Fr{\'e}d{\'e}ric},
	publisher = {Elsevier},
	title = {Sub-seasonal to seasonal prediction: the gap between weather and climate forecasting},
	year = {2018}}

@article{Yan2025,
	author = {Yan, Lin},
	journal = {IEEE Computer Graphics and Applications},
	number = {4},
	pages = {89--98},
	publisher = {IEEE},
	title = {Topology-Based Visualization Techniques for Scientific Data Exploration},
	volume = {45},
	year = {2025}}

@inproceedings{YanMasoodSridharamurthy2021,
	author = {Yan, Lin and Masood, Talha Bin and Sridharamurthy, Raghavendra and Rasheed, Farhan and Natarajan, Vijay and Hotz, Ingrid and Wang, Bei},
	booktitle = {Computer Graphics Forum},
	number = {3},
	organization = {Wiley Online Library},
	pages = {599--633},
	title = {Scalar field comparison with topological descriptors: Properties and applications for scientific visualization},
	volume = {40},
	year = {2021}}

@article{VitartRobertson2018,
	author = {Vitart, Fr{\'e}d{\'e}ric and Robertson, Andrew W},
	journal = {npj climate and atmospheric science},
	number = {1},
	pages = {3},
	publisher = {Nature Publishing Group UK London},
	title = {The sub-seasonal to seasonal prediction project (S2S) and the prediction of extreme events},
	volume = {1},
	year = {2018}}

@article{RobertsonVitartCamargo2020,
	author = {Robertson, Andrew W and Vitart, Frederic and Camargo, Suzana J},
	journal = {Journal of Geophysical Research: Atmospheres},
	number = {6},
	pages = {e2018JD029375},
	publisher = {Wiley Online Library},
	title = {Subseasonal to seasonal prediction of weather to climate with application to tropical cyclones},
	volume = {125},
	year = {2020}}

@article{LamSanchez-GonzalezWillson2023,
	author = {Lam, Remi and Sanchez-Gonzalez, Alvaro and Willson, Matthew and Wirnsberger, Peter and Fortunato, Meire and Alet, Ferran and Ravuri, Suman and Ewalds, Timo and Eaton-Rosen, Zach and Hu, Weihua and others},
	journal = {Science},
	number = {6677},
	pages = {1416--1421},
	publisher = {American Association for the Advancement of Science},
	title = {Learning skillful medium-range global weather forecasting},
	volume = {382},
	year = {2023}}

@article{BiXieZhang2023,
	author = {Bi, Kaifeng and Xie, Lingxi and Zhang, Hengheng and Chen, Xin and Gu, Xiaotao and Tian, Qi},
	journal = {Nature},
	number = {7970},
	pages = {533--538},
	publisher = {Nature Publishing Group UK London},
	title = {Accurate medium-range global weather forecasting with 3D neural networks},
	volume = {619},
	year = {2023}}

@article{ChenZhongLi2024,
	author = {Chen, Lei and Zhong, Xiaohui and Li, Hao and Wu, Jie and Lu, Bo and Chen, Deliang and Xie, Shang-Ping and Wu, Libo and Chao, Qingchen and Lin, Chensen and others},
	journal = {Nature Communications},
	number = {1},
	pages = {6425},
	publisher = {Nature Publishing Group UK London},
	title = {A machine learning model that outperforms conventional global subseasonal forecast models},
	volume = {15},
	year = {2024}}

@article{MouatadidOrensteinFlaspohler2023,
	author = {Mouatadid, Soukayna and Orenstein, Paulo and Flaspohler, Genevieve and Cohen, Judah and Oprescu, Miruna and Fraenkel, Ernest and Mackey, Lester},
	journal = {Nature Communications},
	number = {1},
	pages = {3482},
	publisher = {Nature Publishing Group UK London},
	title = {Adaptive bias correction for improved subseasonal forecasting},
	volume = {14},
	year = {2023}}

@inproceedings{HatanpaaKuStock2025,
	author = {Hatanp{\"a}{\"a}, V{\"a}in{\"o} and Ku, Eugene and Stock, Jason and Emani, Murali and Foreman, Sam and Jung, Chunyong and Madireddy, Sandeep and Nguyen, Tung and Sastry, Varuni and Sinurat, Ray AO and others},
	booktitle = {Proceedings of the International Conference for High Performance Computing, Networking, Storage and Analysis},
	pages = {72--85},
	title = {Aeris: Argonne earth systems model for reliable and skillful predictions},
	year = {2025}}

\end{document}


\maketitle

\section*{S1. Data Preprocessing and Input Construction}

This section describes all preprocessing and input construction steps performed prior to model training, including data normalization,
structural channel derivation, and the construction of dual-trend temporal inputs.

\subsection*{Data source and preprocessing}

We use the ERA5 reanalysis dataset provided by the European Centre for Medium-Range Weather Forecasts (ECMWF)~\cite{HersbachBellBerrisford2020}. The analysis focuses on daily mean 2-m air temperature (T2m) over the continental United States (CONUS). ERA5 fields are sampled on a $0.25^\circ \times 0.25^\circ$
latitude--longitude grid and cropped to a fixed spatial domain of $101 \times 237$ grid points covering the CONUS region.

Hourly ERA5 T2m fields are aggregated into daily means prior to all experiments. Each daily record is treated as a two-dimensional scalar
field. All input channels are normalized to the $[0,1]$ range using the 1st--99th percentile values computed from the training data.

The dataset spans the period 2010--2024. Data from 2010--2019 are used for training and validation, while the period 2020--2024 is reserved as
an independent test set. All reported results are computed on this held-out evaluation period.

\subsection*{Structural input representation}

Each daily temperature field is represented as a multi-channel tensor
\[
\X_t = [\SF_t, \T_t, \V_t, \C_t],
\]
where $\SF_t$ denotes the normalized 2-m air temperature field at time $t$. The auxiliary channels $\T_t$ and $\V_t$ encode the type (local maxima, minima, and saddles) and scalar value of critical points identified in the temperature field, respectively. The channel $\C_t$ represents
saddle-level contour structures extracted from the same field. These channels provide an explicit structural summary of the temperature field, highlighting mesoscale organization that is not captured by raw values alone.

\begin{figure}[h]
\centering
\includegraphics[width=0.6\textwidth]{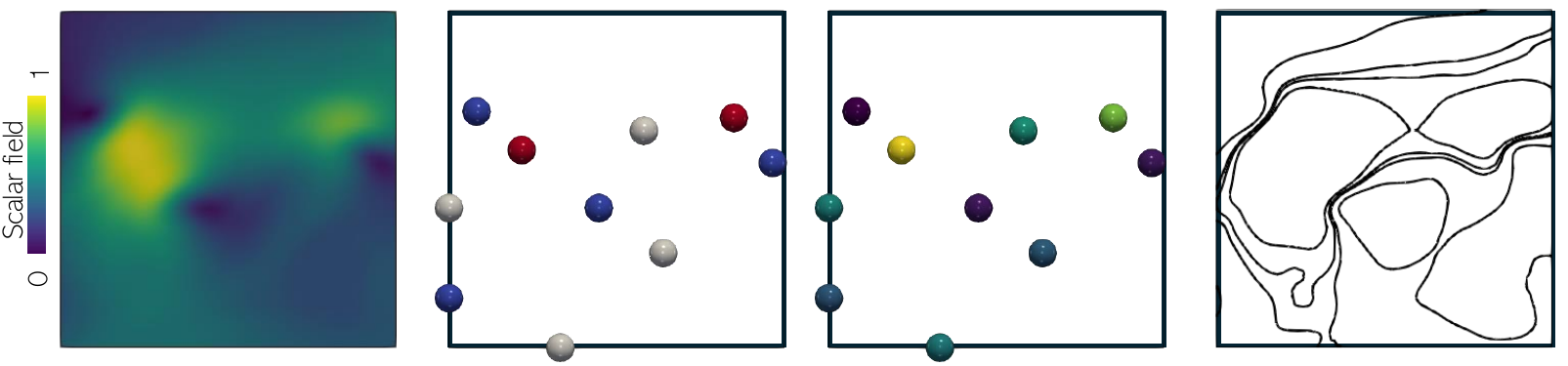}
\caption{Example input channels derived from a toy scalar field for illustration. Shown are the 2D scalar field $\SF$, the critical-point type map $\T$  (red: local maxima, blue: local minima, gray: saddles), the critical-point value map $\V$, and the saddle-level contour map $\C$.}
\label{fig:input}
\end{figure}

All channels share the same spatial resolution ($101 \times 237$ grid points) and are derived solely from the temperature field at time $t$, without incorporating information from future dates. Figure~\ref{fig:input} provides an illustrative example of the four input channels derived from a toy scalar field for visualization purposes.

\subsection*{Dual-trend temporal input construction}

For each target date $t$, two complementary temporal input groups are constructed to capture distinct sources of S2S temperature variability.

\paragraph{Interannual (calendar-aligned) context.}
The interannual input is constructed using temperature fields from the same calendar date in previous years:
\[
\X^{\mathrm{inter}}_t =
\{\X_{t-3\times365},\; \X_{t-2\times365},\; \X_{t-365}\},
\]
with leap years handled by replacing 365 with 366 where appropriate. This input encodes calendar-aligned background variability using
historical observations only.

\paragraph{(Lead-time–matched) recent dynamics.}
The recent input captures short-term temporal evolution preceding the forecast initialization. For each training sample, a forecast lead
time $\tau$ is sampled uniformly from the range 30--90 days, and the recent input is constructed as
\[
\X^{\mathrm{intra}}_t =
\{\X_{t-3\tau},\; \X_{t-2\tau},\; \X_{t-\tau}\}.
\]
This horizon-conditioned grouping encodes recent temperature progression at a temporal scale matched to the forecast lead time.
This design avoids mixing incompatible temporal scales across short and long lead forecasts.

\subsection*{Temporal independence and evaluation protocol}

All experiments adopt strictly year-disjoint temporal splits to prevent information leakage across training and evaluation periods. The dataset spans 2010–2024, with 2010–2019 used for training and validation, and 2020–2024 reserved as an independent test set.

This year-level separation ensures that the model does not access future seasonal realizations when learning calendar-aligned patterns. In particular, interannual context inputs are constructed exclusively from years within the training period, and normalization statistics are computed using training data only. 

These constraints ensure that all evaluation results reflect genuine out-of-sample forecasting behavior rather than information leakage from future observations.

\section*{S2. Model Architecture}

\begin{figure}[h]
\centering
\includegraphics[width=0.75\textwidth]{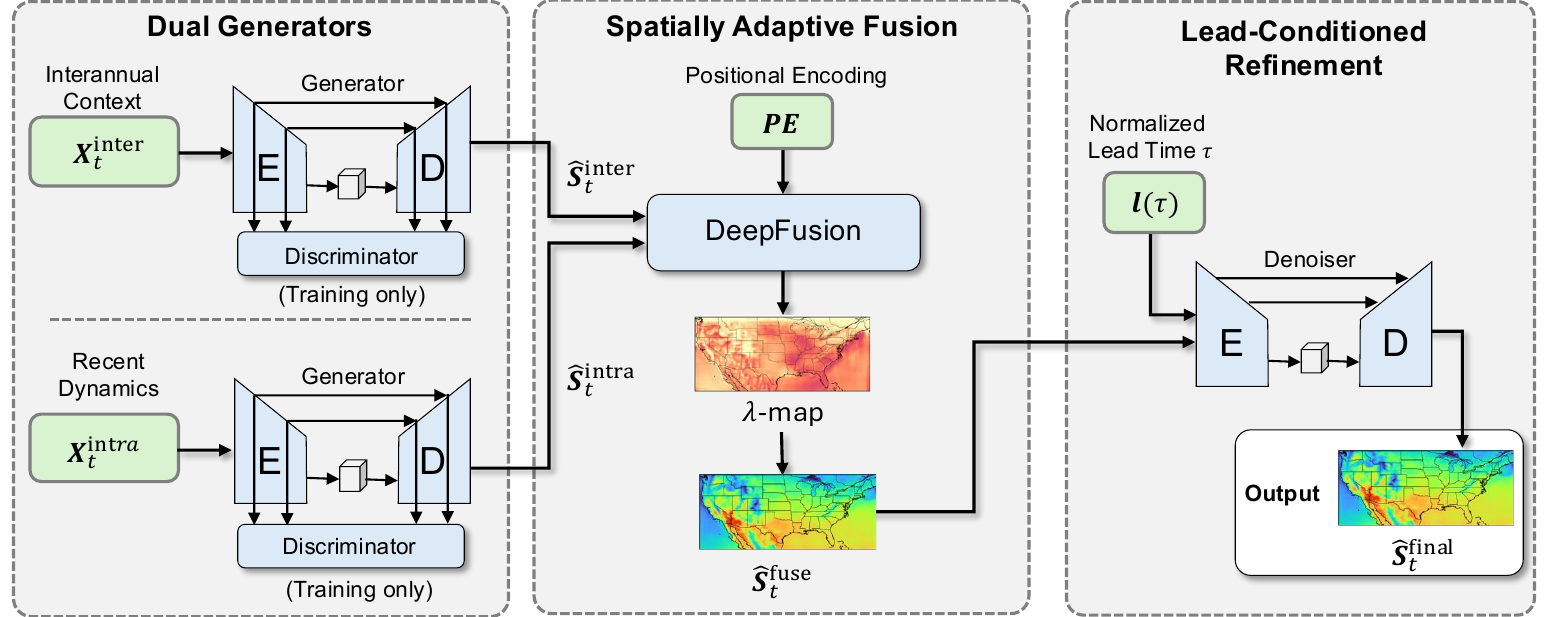}
\caption{Model architecture overview.
The model consists of three sequential modules.
Left: Two parallel conditional generators (InterGen and IntraGen), each implemented as a U-Net--style encoder--decoder, produce initial
temperature predictions from interannual and lead-time–matched input groups, respectively. Each generator is paired with an auxiliary discriminator used only during training.
Center: The two predictions are combined by the spatially adaptive fusion module (DeepFusion). The predictions are concatenated
with a two-channel positional encoding (latitude and longitude) and passed to a lightweight fusion encoder that predicts a spatially varying
$\lambda$-map. The $\lambda$-map controls the local blending of the two predictions and is followed by a residual correction to produce the fused field.
Right: The fused prediction is further refined by a lead-conditioned refinement module implemented as a U-Net denoiser. The refinement network is conditioned on a non-learned lead-time map $\ell(\tau)$, obtained by normalizing the forecast lead time $\tau$ and broadcasting it to the spatial grid, and outputs the final temperature forecast.}
\label{fig:ML-overview}
\end{figure}

The model contains three components that operate in sequence:
(i) two generative predictors produce initial temperature forecasts from two temporal input groups,
(ii) a spatially adaptive fusion module combines the two predictions with spatially varying weights and applies a residual local adjustment, and
(iii) a lead-conditioned refinement module improves the fused forecast by reducing noise and increasing spatial coherence (Fig.~\ref{fig:ML-overview}).

\subsection*{Dual Generators}

The interannual input group $X^{\mathrm{inter}}_t$ and the lead-time–matched input group $X^{\mathrm{intra}}_t$ (Section~S1) are processed by two
separate conditional generators, \textit{InterGen} and \textit{IntraGen}. For a target date $t$, each input group consists of three four-channel
tensors, and each generator produces an initial estimate of the target temperature field:
\[
\hat{\SF}^{\mathrm{inter}}_t=\mathrm{InterGen}(\Xinter_t),
\qquad
\hat{\SF}^{\mathrm{intra}}_t=\mathrm{IntraGen}(\Xintra_t).
\]
Each output is a $101 \times 237$ temperature map scaled to $[0,1]$. Both generators use a U-Net--style encoder--decoder architecture with skip connections.

Each generator is paired with an auxiliary discriminator that evaluates consistency between the generated prediction and the corresponding
ground-truth temperature field $\SF_t$ (contained in $\X_t$). The discriminators are used only during training; inference uses the generators and the downstream fusion/refinement modules.

\subsection*{Spatially Adaptive Fusion (DeepFusion)}

DeepFusion combines $\hat{\SF}^{\mathrm{inter}}_t$ and $\hat{\SF}^{\mathrm{intra}}_t$ through a spatially varying fusion weight map $\LM_t$ (\textit{$\lambda$-map}). To provide explicit spatial context, the two predictions are augmented with a two-channel positional encoding  $\mathbf{PE}$ consisting of normalized latitude and longitude coordinates. The augmented input is
\[
\mathbf{Z}_t = \bigl[
\hat{\SF}^{\mathrm{inter}}_t,\;
\hat{\SF}^{\mathrm{intra}}_t,\;
\mathbf{PE}
\bigr].
\]
A lightweight convolutional encoder extracts joint spatial features $\mathbf{E}(\mathbf{Z}_t)$, from which DeepFusion predicts multi-resolution fusion maps:
\[
\LM^i_t = \sigma\!\left(W_i * \mathbf{E}(\mathbf{Z}_t) + b_i\right),
\qquad i \in \{1,2,3\},
\]
where $\sigma(\cdot)$ is the sigmoid function and $*$ denotes convolution. The finest-resolution map $\LM_t^1(x,y)$ determines the local blending ratio. The fused field is computed as
\[
\SF_t^{\mathrm{fuse}}(x,y)
= \LM_t^1(x,y)\,\hat{\SF}^{\mathrm{inter}}_t(x,y)
  + \bigl(1-\LM_t^1(x,y)\bigr)\,\hat{\SF}^{\mathrm{intra}}_t(x,y).
\]
DeepFusion further applies a residual correction term:
\[
\hat{\SF}_t^{\mathrm{fuse}}(x,y)
= \SF_t^{\mathrm{fuse}}(x,y) + \Delta\SF_t(x,y),
\]
where $\Delta\SF_t(x,y)$ is predicted by the fusion network to provide local adjustments while preserving the global structure induced by the weighted combination.

\subsection*{Lead-Conditioned Refinement Module}

The fused prediction $\hat{\SF}_t^{\mathrm{fuse}}$ is passed to a lead-conditioned refinement module implemented as a U-Net. The module operates on $\hat{\SF}_t^{\mathrm{fuse}}$ and incorporates the forecast lead time through an explicit, non-learned conditioning signal.
Specifically, the forecast lead time $\tau$ is normalized to the range $[0,1]$ and broadcast to a spatial grid to form a constant conditioning
map, denoted by $\boldsymbol{\ell}(\tau)$. This lead-time map is concatenated with the fused field and injected into intermediate layers
of the U-Net to modulate the refinement behavior across different forecast horizons.

The final temperature forecast is predicted as
\begin{equation}
\hat{\SF}_t^{\mathrm{final}}
= F_{\theta}\bigl(\hat{\SF}_t^{\mathrm{fuse}},\, \boldsymbol{\ell}(\tau)\bigr),
\end{equation}
where $F_{\theta}$ denotes the refinement network. The refinement stage is deterministic and performs a single learned update rather than an iterative or stochastic diffusion process.

\section*{S3. Training Strategy and Loss Functions}

The model is trained using a unified optimization procedure that jointly updates the dual generators, the DeepFusion module, and the lead-conditioned refinement network. This design allows the system to integrate multi–time-scale information
end-to-end, while auxiliary constraints are introduced gradually to stabilize training.

The overall objective combines several complementary loss terms:
\begin{equation}
  L
  = \alpha \, L_{\text{content}}
  + \beta \, L_{\text{adv}}
  + \gamma \, L_{\text{reg}}
  + \delta \, L_{\text{topo}},
  \label{eq:generator-loss}
\end{equation}
where \(L_{\text{content}}\) enforces pixel-level and structural fidelity, \(L_{\text{adv}}\) encourages realism through adversarial training, \(L_{\text{reg}}\) regularizes the fusion weights, and \(L_{\text{topo}}\) maintains topological consistency.

\paragraph{Content loss.}
The content loss enforces pixel-level accuracy and structural fidelity of the predicted temperature field:
\begin{equation}
  L_{\text{content}}
  = \operatorname{MAE}\!\bigl(\hat{\SF}_t^{\text{final}}, \SF_t\bigr)
  + \Bigl(1 - \operatorname{SSIM}\!\bigl(\hat{\SF}_t^{\text{final}}, \SF_t\bigr)\Bigr),
\end{equation}
where MAE measures pointwise error and SSIM~\cite{WangBovikSheikh2004} captures structural similarity between two-dimensional fields.
\paragraph{Adversarial loss.}
Adversarial training is used to encourage the realism of predicted temperature fields.
We adopt a hinge GAN formulation~\cite{MiyatoKataokaKoyama2018}. The discriminator loss is
\begin{equation}
  L_D
  = \mathbb{E}\Bigl[\max\bigl(0,\, 1 - D(\X_t)\bigr)\Bigr]
  + \mathbb{E}\Bigl[\max\bigl(0,\, 1 + D(\hat{\X}_t^{\text{fake}})\bigr)\Bigr],
\end{equation}
and the corresponding adversarial loss for the generator is
\begin{equation}
  L_{\text{adv}}
  = - \mathbb{E}\bigl[D(\hat{\X}_t^{\text{fake}})\bigr].
\end{equation}

where \(D(\cdot)\) denotes the discriminator, $\X_t$ are real samples (concatenated scalar and auxiliary maps), and $\hat{\X}_t^{\text{fake}}$ are generated samples for each branch and for the fused output.

\paragraph{Fusion regularization.}
To avoid degenerate spatial fusion behavior, the finest-resolution fusion map $\LM_t^1$ is regularized using a combination of smoothness, entropy, and mean-balance terms:
\begin{equation}
  L_{\text{reg}}
  = \eta_1 \, \TV(\LM_t^1)
  + \eta_2 \, H(\LM_t^1)
  + \eta_3 \bigl(\overline{\LM_t^1} - \lambda_{\text{target}}\bigr)^2.
\end{equation}
These terms encourage spatially smooth and non-saturated fusion weights without enforcing a fixed blending pattern.

\paragraph{Topological loss.}
To further encourage the structural consistency of the predicted scalar fields, we include an auxiliary topological loss based on persistent homology~\cite{EdelsbrunnerHarer2010}:
\begin{equation}
  L_{\text{topo}}
  = d_B\Bigl(
    \operatorname{PD}(\SF_t),
    \operatorname{PD}(\hat{\SF}_t^{\text{final}})
  \Bigr),
\end{equation}
where $\operatorname{PD}(\cdot)$ denotes the persistence diagram and $d_B$ is the bottleneck distance, computed using the Gudhi library~\cite{Project2025}. We focus on one-dimensional homology ($H_1$) features.

To reduce computational overhead and prevent interference with early training dynamics, the topological loss is activated only after an initial warm-up period and evaluated intermittently (every five
mini-batches).
As a result, $L_{\text{topo}}$ serves as a late-stage auxiliary regularizer rather than a primary optimization objective.

\section*{S4. Additional Experimental Analyses}

Forecast skill is quantified using RMSE, PSNR, SSIM, and ACC, following standard definitions in subseasonal-to-seasonal climate forecasting~\cite{VitartArdilouzeBonet2017}.

\subsection*{S4.1 Qualitative Comparison with the Swift Baseline}
To complement the quantitative comparison with the Swift baseline shown in Fig. 2C, we provide a representative spatial comparison across lead times.

\begin{figure*}[h!]
\centering
\includegraphics[scale=0.6]{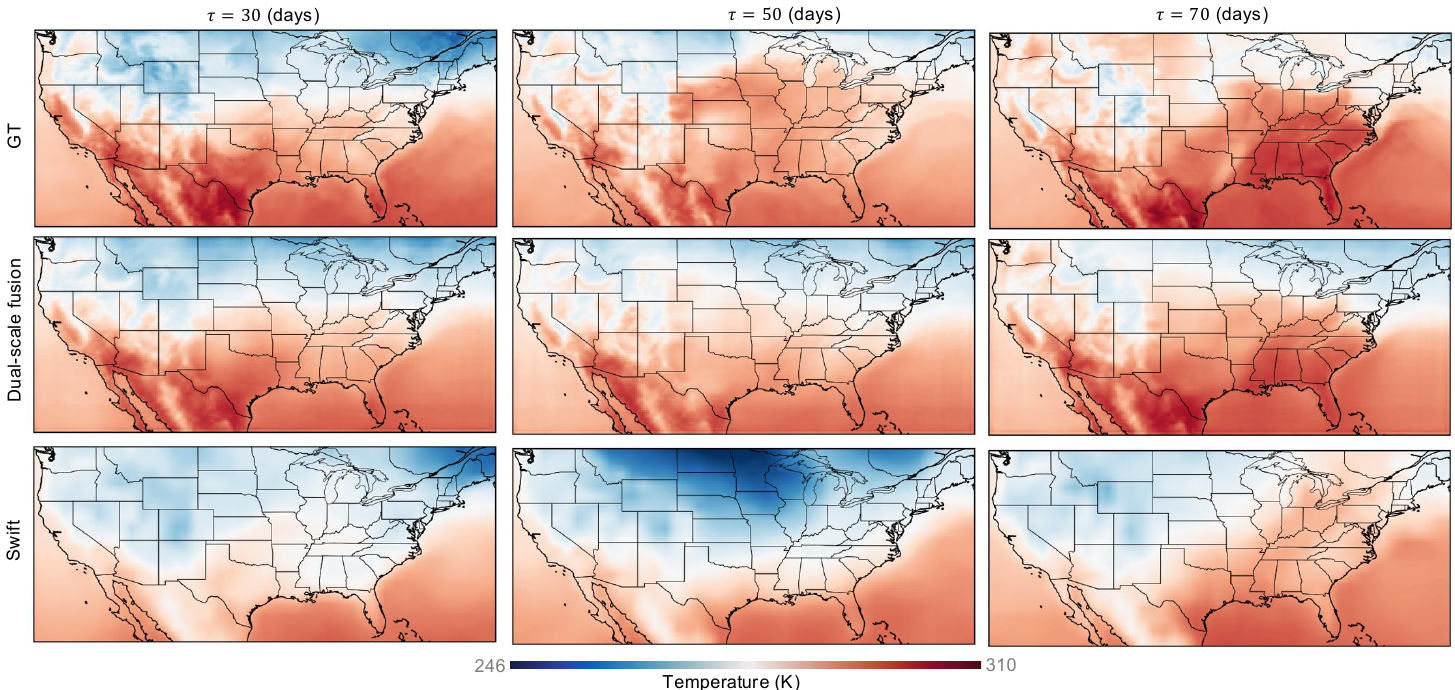}
\caption{
Qualitative comparison with the Swift baseline across lead times. Ground truth (GT), dual-scale fusion prediction, and Swift prediction are shown for a case initialized on January 19, 2020, evaluated at lead times of 30, 50, and 70 days. The example is randomly selected from the test period. While both methods exhibit comparable error levels, the dual-scale model maintains more coherent spatial structure across horizons, whereas Swift predictions show reduced structural consistency at longer lead times.
}
\label{fig:Swift}
\end{figure*}

Figure~\ref{fig:Swift} presents forecasts initialized on January 19, 2020, evaluated at lead times of 30, 50, and 70 days. This case is randomly selected from the test period and is not chosen based on model performance.

Consistent with the aggregate results in Fig. 2C, both methods achieve comparable error levels within the overlapping lead-time range. However, the dual-scale model preserves more coherent large-scale spatial structures across horizons, whereas the Swift predictions exhibit reduced structural consistency, particularly at longer lead times.

\subsection*{S4.2 Seasonal Structure Across the Annual Cycle}
\begin{figure*}[h!]
\centering
\includegraphics[scale=0.5]{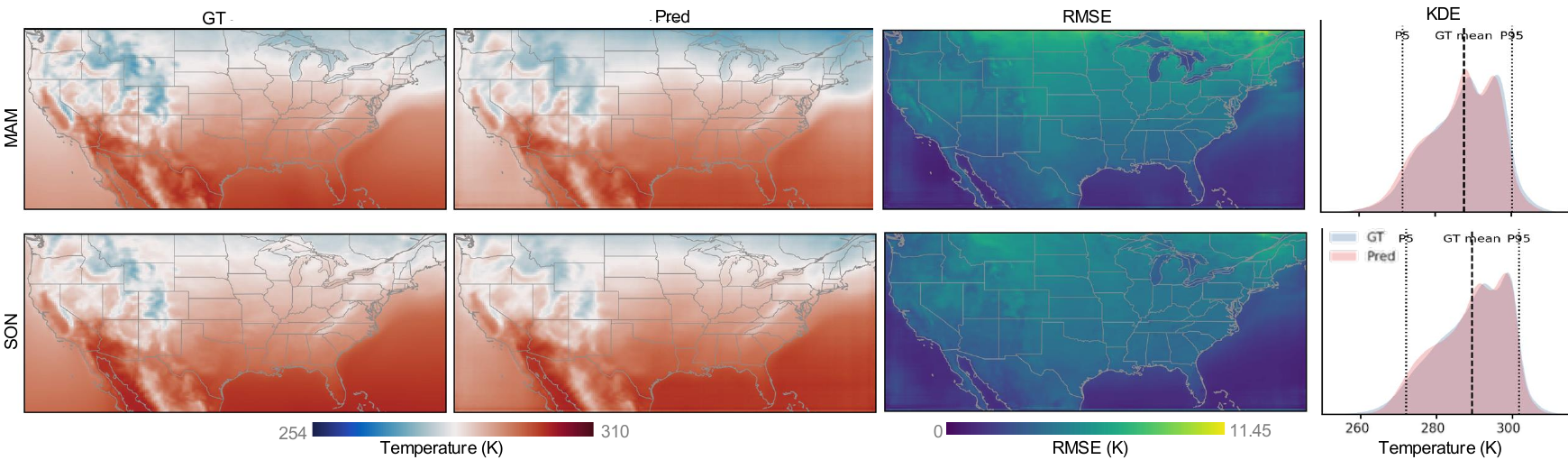}
\caption{
Forecast performance during transitional seasons.
Top row: March–April–May (MAM).
Bottom row: September–October–November (SON).
From left to right: seasonal mean ground truth temperature, model prediction, spatial RMSE, and corresponding kernel density estimate (KDE) comparison between prediction and ground truth distributions.
Both transitional seasons exhibit intermediate predictability relative to winter (DJF) and summer (JJA), with spatial error patterns consistent with continental interior amplification.
The close agreement between predicted and observed temperature distributions further indicates that the model maintains distributional fidelity across seasonal regimes.
}
\label{fig:Seasonal}
\end{figure*}

To extend the seasonal analysis beyond the winter (DJF) and summer (JJA) cases shown in the main text, we examine forecast behavior across the full annual cycle, including the transitional seasons spring (MAM) and autumn (SON). Figure~\ref{fig:Seasonal} presents seasonal mean ground truth, prediction, spatial RMSE, and distributional comparisons for MAM and SON. Table~\ref{tab:seasonal_stats} summarizes seasonal forecast statistics across all four seasons, while Table~\ref{tab:seasonal_conrast} reports seasonal contrast and lead-time sensitivity relative to a simple day-of-year climatology baseline.

\begin{table}[h!]
\centering
\caption{Seasonal forecast performance statistics across all four seasons. 
Mean RMSE (K), standard deviation (Std), and anomaly correlation coefficient (ACC) are computed over the independent 2020–2024 test period and averaged across lead times from 30–90 days (5-day intervals). 
ACC is calculated using temperature anomalies relative to the training-period calendar-day climatology.}
\label{tab:seasonal_stats}
\begin{tabular}{lcccc}
\toprule
Season & Mean RMSE (K) & Std (K) & ACC & Overlap \\
\midrule
DJF & 5.778 & 1.805 & 0.239 & 0.925 \\
MAM & 4.296 & 1.152 & 0.504 & 0.960 \\
JJA & 2.737 & 0.598 & 0.661& 0.954 \\
SON & 3.865 & 1.133 & 0.577 & 0.970 \\
\bottomrule
\end{tabular}
\end{table}

\begin{table}[t]
\centering
\caption{Seasonal contrast and lead-time sensitivity relative to a day-of-year climatology baseline.}
\label{tab:seasonal_conrast}

\setlength{\tabcolsep}{4pt}
\small

\begin{tabular}{lcccc}
\hline
Model 
& RMSE 
& RMSE 
& $\Delta_{\text{season}}$ 
& Sensitivity \\
& (DJF) & (JJA) & (DJF$-$JJA) & (90$-$30) \\
\hline
Day-of-year climatology & 8.033 & 6.500 & 1.533 & -0.013 \\
Ours (dual-scale)       & 5.778 & 2.737 & 3.041 & 0.348 \\
$\Delta$ (Ours - Climatology) & -2.255 & -3.763 & +1.508 & +0.361 \\
\hline
\end{tabular}
\end{table}

The results confirm a clear seasonal ordering in forecast difficulty. Winter (DJF) exhibits the largest mean RMSE (5.78 K), followed by MAM (4.30 K), SON (3.86 K), and summer (JJA) with the lowest mean RMSE (2.74 K). ACC remains moderate to high across seasons, with the weakest anomaly correlation in winter and the strongest in summer, consistent with the RMSE-based ordering of predictability. Distributional overlap remains high in all cases (> 0.92), indicating that the model preserves the large-scale statistical structure of seasonal temperature fields.

The transitional seasons occupy an intermediate regime between winter and summer in both error magnitude and spatial organization. In parallel, Table~\ref{tab:seasonal_conrast} shows that the day-of-year climatology baseline already exhibits seasonal contrast but negligible lead-time sensitivity, whereas the dual-scale model substantially reduces RMSE while retaining a stronger winter--summer separation. Together, these results show that the seasonal organization of forecast skill extends smoothly across the annual cycle and remains more prominent than the comparatively modest variation across lead times.

\subsection*{S4.3 Lead-Time Dependence of Forecast Skill and Fusion Behavior}

\begin{figure*}[h!]
\centering
\includegraphics[scale=0.8]{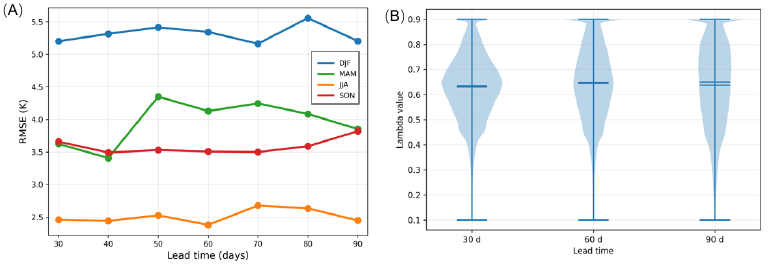}
\caption{
Lead-time dependence of forecast skill and fusion behavior.
(A) Seasonal mean RMSE as a function of lead time. While summer (JJA) consistently exhibits lower RMSE than other seasons, all four seasons display relatively flat lead-time curves, with only modest degradation from 30 to 90 days.
(B) Distribution of the learned fusion weight $\lambda$ across lead times. Median values remain stable across horizons, indicating limited sensitivity of the fusion mechanism to forecast lead time. A slight broadening of the upper tail at longer lead times suggests that high-$\lambda$ regimes become marginally more prevalent as forecast horizon increases.
Overall, these results confirm that seasonal predictability structure dominates over lead time in shaping both forecast skill and fusion behavior. 
}

\label{fig:leadtime}
\end{figure*}

We further examine forecast skill and fusion behavior across lead times 
(30, 60, and 90 days), as shown in Figure~\ref{fig:leadtime}.

Seasonal RMSE curves (Figure~\ref{fig:leadtime}A) exhibit limited variation across the 30–90-day range. Within each season, the spread of RMSE across lead times remains modest compared to the difference between seasonal mean error levels. Summer maintains consistently lower RMSE across all horizons, while winter remains systematically higher; however, none of the seasonal curves display monotonic or rapid error growth with increasing lead time. These patterns indicate that, within the subseasonal window considered here, forecast skill degradation with horizon is comparatively gradual relative to seasonal differences.

The distribution of the learned fusion weight $\lambda$ (Figure~\ref{fig:leadtime}B) remains broadly stable across lead times. Median values at 30, 60, and 90 days are nearly identical, and the overall spread of the distributions shows no systematic shift in central tendency.

A modest increase in upper-tail dispersion is visible at longer lead times, but this change reflects a slight broadening of the distribution rather than a displacement of its median. These patterns indicate that, within the 30–90-day window, lead time exerts limited influence on the central fusion behavior and primarily modulates distributional spread. This further supports the interpretation that seasonal structure and local forecast difficulty play a more prominent role in organizing $\lambda$ than forecast horizon alone.

\subsection*{S4.4 Regime-Level $\lambda$ Behavior Across Seasons}

\begin{figure*}[h!]
\centering
\includegraphics[scale=0.8]{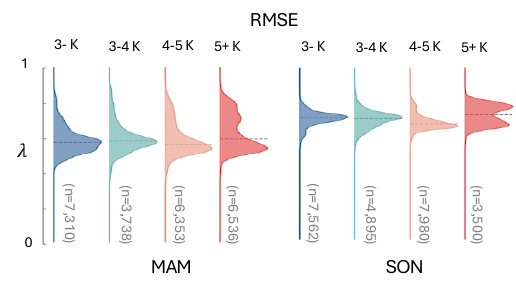}
\caption{
Ridgeline distributions of the learned fusion weight $\lambda$ stratified by seasonal mean RMSE for spring (MAM) and autumn (SON).
Grid points are grouped into four fixed RMSE bins (3--, 3--4, 4--5, and 5+ K). Each ridge represents the kernel density estimate of $\lambda$ within one bin. Dashed lines indicate median values, and $n$ denotes the number of grid points.
Within each season, median $\lambda$ varies only modestly across RMSE bins, with no clear monotonic shift across difficulty regimes.
However, the overall $\lambda$ distribution in SON is shifted toward higher values compared with MAM, indicating season-dependent baseline differences in fusion behavior.
}
\label{fig:ridge}
\end{figure*}

To quantify the regime-level behavior of the learned fusion weight, we stratify grid points by seasonal mean RMSE into four fixed bins (3-, 3--4, 4--5, and 5+ K) and examine the distribution of $\lambda$ within each bin. Figure~\ref{fig:ridge} extends the winter (DJF) and
summer (JJA) analysis shown in Fig.~4D of the main text by including
the transitional seasons spring (MAM) and autumn (SON). Each ridge
represents the kernel density estimate of $\lambda$ within a given
RMSE bin, with dashed lines indicating median values.

For the transitional seasons, both MAM and SON exhibit weak and
non-monotonic variation across RMSE bins, with substantial overlap
in the ridge distributions. In MAM, median $\lambda$ values fluctuate
slightly across bins, while SON shows a mild overall increase with
noticeable deviations from monotonic behavior.

Table~\ref{tab:lambda_median_bins} reports the median $\tilde{\lambda}$
for each season and RMSE bin, together with
$\Delta = \tilde{\lambda}_{5+} - \tilde{\lambda}_{3\text{--}K}$.
The largest $\Delta$ occurs in DJF, whereas MAM and SON show smaller
positive separation and JJA shows near-zero or slightly negative
separation. These results reinforce that $\lambda$ adapts to forecast
difficulty in a strongly season-dependent manner.

\begin{table}[h]
\centering
\caption{
Median fusion weight $\tilde{\lambda}$ stratified by seasonal mean RMSE bins.
The final column reports
$\Delta = \tilde{\lambda}_{5+} - \tilde{\lambda}_{3-}$,
where $\tilde{\lambda}_{b}$ denotes the median $\lambda$ over grid points in RMSE bin $b$. 
}
\label{tab:lambda_median_bins}
\begin{tabular}{lccccc}
\toprule
Season & 3-- K & 3--4 K & 4--5 K & 5+ K & $\Delta$ \\
\midrule
DJF & 0.687 &  0.698&0.737  & 0.742 &  0.055\\
MAM & 0.572 &0.580  &0.562  &0.592  &0.020  \\
JJA &  0.554& 0.497 & 0.509 & 0.517 & -0.037 \\
SON & 0.702 & 0.706 & 0.673 & 0.727 & 0.025 \\
\bottomrule
\end{tabular}
\end{table}

\subsection*{S4.5 Topology Effects on Optimization and Predictive Performance}

We further examine the impact of topological constraints on both optimization dynamics and predictive behavior.

\begin{figure*}[b]
\centering
\includegraphics[scale=1]{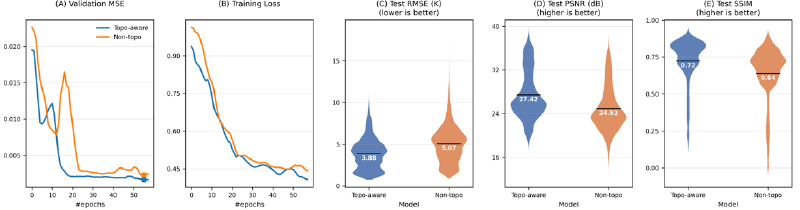}
\caption{
Topology-aware modeling improves optimization stability and predictive consistency.
(A) Validation MSE during training. The topology-aware model reaches the low-error regime earlier and avoids the mid-stage validation spike observed in the matched baseline without topological inputs.
(B) Training loss curves for the two models.
(C–E) Distribution of test-set performance metrics (RMSE, PSNR, and SSIM). The topology-aware formulation shifts the distributions toward lower error and higher structural similarity across evaluation samples.
}
\label{fig:topology}
\end{figure*}

\paragraph{Optimization dynamics.}
Figs.~\ref{fig:topology}A--B compare the validation and training behavior of the topology-aware model and the matched baseline without topological inputs. While both models converge during training, the topology-aware formulation enters a low-error validation regime earlier and avoids the pronounced mid-stage validation spike observed in the baseline. This pattern suggests that topology-aware descriptors do not merely reduce the final error level, but also stabilize the training trajectory. Training loss remains consistently lower as well, although the difference is more modest than in validation.

\paragraph{Distribution of predictive performance.}
Figs.~\ref{fig:topology}C--E show the distribution of test-set RMSE, PSNR, and SSIM across evaluation samples. 
Compared with the baseline, the topology-aware model shifts the distributions toward lower RMSE and higher structural similarity. 
Importantly, the improvement appears as a broad distributional shift rather than being driven by a small number of extreme cases. 
The overall spread of the distributions remains comparable, suggesting that gains arise from systematic improvements across samples rather than variance amplification.

Taken together, these observations complement the spatial error reductions shown in Fig.~5 of the main text. Topology-aware descriptors lead to smoother optimization trajectories and more consistent predictive behavior across evaluation samples, suggesting a stabilizing effect on spatial dependency learning.

\section*{Reproducibility checklist}

\subsection*{Data availability}
The ERA5 reanalysis dataset used in this study is publicly available from the European Centre for Medium-Range Weather Forecasts (ECMWF). Details of data preprocessing, normalization, and input construction are described in Section S1 (Data Preprocessing and Input Construction).

\subsection*{Code availability}
The code used for data preprocessing, model training, evaluation, and figure generation is maintained in the VisTALE Lab GitHub repository (S2STemp), currently hosted at
https://github.com/
VisTALELab/S2STemp.
The repository is currently private during peer review and will be made publicly available upon publication. Access can be provided to editors and reviewers upon request.

\subsection*{Model architecture}
A detailed description of the model architecture, including the dual generators, spatially adaptive fusion module, and the lead-conditioned refinement network, is provided in Section S2 (Model Architecture).

\subsection*{Training procedure}
The training strategy, loss functions, and optimization details are described in Section S3 (Training Strategy and Loss Functions).

\subsection*{Experimental evaluation}
Additional experimental analyses, evaluation metrics, and seasonal and lead-time experiments are described in Section S4 (Additional Experimental Analyses).

\subsection*{Experimental protocol}
The dataset spans 2010–2024, with 2010–2019 used for training and validation and 2020–2024 reserved for testing. The full experimental protocol and evaluation procedure are described in the Materials and Methods section of the main text and the Supporting Information.

\bibliographystyle{naturemag}
\bibliography{SSTemp-SI}